\documentclass[runningheads,a4paper]{llncs}

\usepackage{times}
\usepackage{amssymb}
\usepackage{graphicx}
\usepackage{url}
\usepackage[ruled,vlined]{algorithm2e}
\usepackage{color}
\usepackage{multirow}
\usepackage[T2A,T1]{fontenc}
\usepackage[utf8]{inputenc}
\usepackage[russian,english]{babel}

\newcommand{\keywords}[1]{\par\addvspace\baselineskip
\noindent\keywordname\enspace\ignorespaces#1}

\urldef{\mailsa}\path|louk_nat@mail.ru|
\urldef{\mailsb}\path|mnokel@gmail.com|
\urldef{\mailsc}\path|ivanov.kir.m@yandex.ru
|    

\begin{document}

\title{Combining Thesaurus Knowledge and \\ Probabilistic Topic Models}

\titlerunning{Combining Thesaurus Knowledge and Probabilistic Topic Models}

\author{Natalia Loukachevitch \inst{1} \and Michael Nokel\inst{2} \and Kirill Ivanov\inst{3}}

\authorrunning{Loukachevitch et al.}

\institute{Lomosov Moscow State University, Moscow Russia \\
\mailsa\\
\and
Yandex, Moscow, Russsia \\
\mailsb\\
\and
Lomosov Moscow State University, Moscow Russia \\
\mailsc\\
}

\index{Surname1, Firstname1}
\index{Surname2, Firstname2}

\toctitle{} \tocauthor{}

\maketitle

\begin{abstract}
In this paper we present  the approach of introducing  thesaurus knowledge into probabilistic topic models. The main idea of the approach is based on the assumption that the frequencies of semantically related words and phrases, which are met in the same texts, should be enhanced: this action leads to their larger contribution into topics found in these texts. We have conducted experiments with several thesauri and found that for improving topic models, it is useful to utilize domain-specific knowledge.  If a general thesaurus, such as WordNet, is used, the thesaurus-based improvement of topic models can be achieved with excluding hyponymy relations  in combined topic models.

\end{abstract}

\section{Introduction}

Currently, probabilistic topic models are important tools for improving automatic text processing including information retrieval, text categorization, summarization, etc. Besides, they can be useful  in supporting expert analysis of document collections, news flows, or large volumes of messages in social networks \cite {Blei2012,Smith,Chang}. To facilitate this analysis, such approaches as automatic topic labeling   and various visualization techniques  have been proposed \cite{Smith,Blei2009}. 

        Boyd-Graber et al. \cite {Boyd}  indicate that to be understandable by humans, topics should be specific, coherent, and informative. Relationships between the topic components can be inferred. 
        In \cite {Smith} four topic visualization approaches are compared.  The authors of the experiment concluded that manual topic labels include a considerable number of phrases; users prefer shorter labels with more general words and  tend to incorporate phrases and more generic terminology when using more complex network graph. Blei and Lafferty \cite{Blei2009}   visualize  topics  with  ngrams consisting of words mentioned in these topics. These works show that phrases and knowledge about hyponyms/hypernyms are important for topic representation.

       In this paper we describe an approach  to integrate  large manual lexical resources such as WordNet or EuroVoc into probabilistic topic models, as well as automatically extracted n-grams to improve coherence and informativeness of generated topics. The structure of the paper is as follows. In Section 2 we consider related works. Section 3 describes the proposed approach. Section 4 enumerates  automatic quality measures used in experiments. Section 5 presents the  results obtained on several text collections according to automatic measures. Section 6 describes  the results of manual evaluation of combined topic models for  Islam Internet-site thematic analysis.

\section{Related Work}
Topic modeling approaches are unsupervised statistical algorithms that usually considers each document as a "bag of words".  There were several attempts to enrich word-based topic models (=unigram topic models) with additional prior knowledge or multiword expressions.

 Andrzejewski et al. \cite {Andrzejewski} incorporated knowledge by Must-Link and Cannot-Link primitives represented by a Dirichlet Forest prior. These primitives were then used in \cite {Newman}, where similar words are encouraged to have similar topic distributions. However, all such methods incorporate knowledge in a hard and topic-independent way, which is a simplification since two words that are similar in one topic are not necessarily of equal importance for another topic. 

Xie et al. \cite {Xie} proposed  a  Markov Random Field regularized LDA model (MRF-LDA), which utilizes the external knowledge  to improve the coherence of
topic modeling. Within a document,  if  two  words  are  labeled  as  similar  according to the external knowledge, their latent topic nodes are connected by an undirected edge and a binary potential function is defined to encourage
them  to  share  the  same  topic  label. Distributional similarity of words is calculated beforehand on a large text corpus. 

In  \cite {Chen}, the authors gather so-called lexical relation sets (LR-sets) for word senses described in WordNet. The LR-sets include synonyms, antonyms and adjective-attribute related words. To adapt LR-sets to a specific domain corpus and to remove inappropriate lexical relations, the  correlation matrix for word pairs in each LR-set is calculated. This matrix at the first step is used for filtrating inappropriate senses, then it is used to modify the initial LDA topic model according to the generalized Polya urn model described in \cite {Mimno}. The generalized Polya urn model boosts probabilities of related words in word-topic distributions. 

Gao and Wen \cite {Gao} presented Semantic Similarity-Enhanced Topic Model that accounts for corpus-specific word co-occurrence and  word semantic similarity calculated on WordNet paths between corresponding synsets using the generalized Polya urn model.  They apply their topic model for categorizing short texts. 

All above-mentioned approaches on adding knowledge to topic models are limited to single words. 
Approaches  using ngrams in topic models can be subdivided into two groups. The first group of methods tries
 to create a unified probabilistic model accounting unigrams and phrases. Bigram-based approaches include the Bigram Topic Model \cite {Wallach} and LDA Collocation Model \cite  {Griffiths}. In \cite {Wang}  the Topical N-Gram Model was proposed to  allow the generation of ngrams based on the context. However, all these models are  enough complex and hard to compute on real datasets.

The second group of methods is based on preliminary extraction of ngrams and their further use in topics generation. Initial studies of this approach used only bigrams  \cite {Lau2013,Nokel2015}.  Nokel and Loukachevitch  \cite {Nokel2016} proposed the LDA-SIM algorithm, which integrates top-ranked ngrams and terms of information-retrieval thesauri into topic models (thesaurus relations were not utilized). They create similarity sets of expressions having the same word components and sum up frequencies of similarity set members if they co-occur in the same text.

In this paper we describe the approach to integrate whole manual thesauri into topic models together with multiword expressions. 

\section{Approach to Integration Whole Thesauri into Topic Models}

In our approach we develop the idea of \cite {Nokel2016} that  proposed to construct similarity sets between ngram phrases between each other and single words. Phrases and words are included in the same similarity set if they have the same component word, for example, \textit {weapon – nuclear weapon – weapon of mass destruction; discrimination –   racial  discrimination}.  It was supposed that if expressions from the same similarity set co-occur in the same document then their contribution into the document's topics is really more than  it is presented with their  frequencies, therefore their frequencies should be increased. In such an approach, the algorithm can "see" similarities between different multiword expressions with the same component word.

In our approach, at first, we include related single words and phrases from a thesaurus such as WordNet or EuroVoc  in these similarity sets. Then, we add  preliminarily extracted ngrams into these sets and, this way, we use two different sources of external knowledge. We use the same LDA-SIM algorithm as described in \cite {Nokel2016} but study what types of semantic relations can be introduced into such similarity sets and be useful for improving topic models. The pseudocode of LDA-SIM algorithm is presented in Algorithm~\ref{algo:lda_sim_algorithm}, where $S=\{S_w\}$ is a similarity set, expressions in similarity sets can comprise single words, thesaurus phrases  or generated noun compounds.

We can compare this approach with the approaches applying the generalized Polya urn model \cite {Chen,Mimno,Gao}. To add prior knowledge, those approaches change topic distributions for related words globally in the collection. We modify topic probabilities  for related words and phrases locally, in specific texts, only when related words (phrases) co-occur in these texts.

\begin{algorithm}[ht!]
\KwIn{collection $D$, vocabulary $W$, number of topics $|T|$, initial $\{p(w|t)\}$ and $\{p(t|d)\}$, sets of similar expressions $S$, hyperparameters $\{\alpha_t\}$ and $\{\beta_w\}$, $n_{dw}$ is the frequency of $w$ in the document $d$}
\KwOut{distributions $\{p(w|t)\}$ and $\{p(t|d)\}$}
\nl\While{not meet the stop criterion}{
        \nl\For{$d\in D,w\in W,t\in T$}{
                \nl{$p(t|d,w)=\frac{p(w|t)p(t|d)}{\sum\limits_{u\in T}p(w|u)p(u|d)}$}\\
        }
        \nl\For{$d\in D,w\in W,t\in T$}{
                \nl{$\textcolor{red}{n'_{dw}=n_{dw}+\sum\limits_{s\in S_w}n_{ds}}$}\\
                \nl{$p(w|t)=\frac{\sum\limits_{d\in D}\textcolor{red}{n'_{dw}}p(t|d,w)+\beta_w}{\sum\limits_{d\in D}\sum\limits_{w\in d}\textcolor{red}{n'_{dw}}p(t|d,w)+\sum\limits_{w\in W}\beta_w}$}\\
                \nl{$p(t|d)=\frac{\sum\limits_{w\in d}\textcolor{red}{n'_{dw}}p(t|d,w)+\alpha_t}{\sum\limits_{w\in W}\sum\limits_{t\in T}\textcolor{red}{n'_{dw}}p(t|d,w)+\sum\limits_{t\in T}\alpha_t}$}\\
        }
}
\caption{\label{algo:lda_sim_algorithm}LDA-SIM algorithm}
\end{algorithm}

\section{Automatic Measures to Estimate the Quality of Topic Models}
        To estimate the quality of topic models, we use two main automatic measures: topic coherence and kernel uniqueness. For human content analysis, measures of topic coherence and kernel uniqueness are both important and complement each other. Topics can be coherent but have a lot of repetitions. On the other hand, generated topics can be very diverse, but incoherent within each topic.
                
        Topic coherence is an automatic metric of interpretability. It was shown that the coherence measure has a high correlation with the expert estimates of topic interpretability \cite{Mimno,Lau2014}. Mimno \cite{Mimno} described an experiment comparing expert evaluation of LDA-generated topics and automatic topic coherence measures.  It was found that most "bad" topics consisted of words without clear relations between each other. 

        Newman et al. \cite{Newman} asked users to score topics on a 3-point scale, where 3=“useful” (coherent) and 1=“useless” (less coherent). They instructed the users that one indicator of usefulness is the ease by which one could think of a short label to describe a topic. Then several automatic measures, including WordNet-based measures and corpus co-occurrence measures, were compared. It was found that the best automatic measure having the largest correlation with human evaluation is word co-occurrence calculated as point-wise mutual information (PMI) on Wikipedia articles. Later Lau et al. \cite{Lau2014} showed that normalized poinwise mutual information (NPMI)  \cite{Bouma} calculated on Wikipedia articles  correlates even more strongly with human scores.

 We calculate automatic topic coherence using two measure  variants. The coherence of a topic is the median PMI (NPMI) of word pairs  representing the topic, usually it is calculated for $n$ most probable elements (in our study ten elements) in the topic. The coherence of the model is the median of the topic coherence. To make this measure  more objective, it should be calculated on an external corpus \cite {Lau2014}. In our case, we use Wikipedia dumps.

        \begin{equation}
        PMI(w_i, w_j) = log\frac{p(w_i, w_j)}{p(w_i)p(w_j)} \;\;\;\;\;\;\;\;\; NPMI(w_i, w_j) = \frac{PMI(w_i, w_j)}{-log(p(w_i, w_j))}
        \end{equation}

        Human-constructed topics usually have unique main words. The measure of kernel uniqueness shows to what extent topics are different from each other and is calculated as the  number of unique elements among most probable elements of topics (kernels)  in relation to the whole number of elements in kernels.
    
    \begin{equation}
        U(\Phi) = \frac{|\cup_{t}kernel(T_i)|}{\sum_{t \in T} |kernel(T_i)|}
        \label{uk}
        \end{equation}
        
         If uniqueness of the topic kernels is closer to zero then many topics are similar to each other, contain the same words in their kernels. In this paper the kernel of a topic means the ten most probable words in the topic. We also calculated perplexity as the measure of language models. We use it for additional checking the model quality.

\section{Use of Automatic Measures to Assess Combined Models}

 For evaluating topics with automatic quality measures, we used several English text collections and one Russian collection (Table~\ref{tab:text_corpora}). We experiment with three thesauri:  WordNet\footnote {https://wordnet.princeton.edu/} (155 thousand entries), information-retrieval thesaurus of the European Union EuroVoc  (15161 terms)\footnote {http://eurovoc.europa.eu/drupal/}, and Russian thesaurus RuThes (115 thousand entries) \footnote{http://www.labinform.ru/pub/ruthes/index\_eng.htm}\cite {Loukachevitch}.

\begin{table}
\begin{center}
\begin{tabular}{|c|c|c|}
\hline
\textbf{Text collection} & \textbf{Number of texts} & \textbf{Number of words} \\
\hline

\textit{English part of} & \multirow{2}{*}{9672} & \multirow{2}{*}{$\approx$ 56 mln}\\
\textit{Europarl corpus} & & \\
\hline
\textit{English part of} & \multirow{2}{*}{23545} & \multirow{2}{*}{$\approx$ 53 mln}\\
\textit{JRC-Acquiz corpus} & & \\
\hline
\textit{ACL Anthology} & \multirow{2}{*}{10921} & \multirow{2}{*}{$\approx$ 48 mln}\\
\textit{Reference corpus} & & \\
\hline
\textit{NIPS Conference} & \multirow{2}{*}{17400} & \multirow{2}{*}{$\approx$ 5 mln}\\
\textit{Papers (2000--2012)} & & \\
\hline
\textit{Russian banking texts} & 10422 & $\approx$ 32 mln\\
\hline
\end{tabular}

\caption{\label{tab:text_corpora}Text collections for experiments}
\end{center}
\end{table}

At the preprocessing step, documents were processed by morphological analyzers.  Also, we extracted noun groups as described in \cite {Nokel2016}. As baselines, we use the unigram LDA topic model and LDA topic model with added 1000 ngrams with maximal NC-value ~\cite{Frantzi} extracted from the collection under analysis.

 As it was found before \cite {Lau2013,Nokel2016},  the addition of ngrams without accounting relations between their components considerably worsens the perplexity because of the vocabulary growth (for perplexity the less is the better) and practically does not change other automatic quality measures (Table 2).

\begin{table}[h]

\begin{center}
\begin{tabular}{|c|l|c|c|c|c|}
\hline 
\bf Collection& \bf Method&\bf TC-PMI&\bf TC-NPMI & \bf Kernel Uniq&\bf Perplex.\\
    \hline

Europarl&LDA unigram&1.20&0.24&0.33&1466\\
&LDA+1000ngram&1.19&0.23&0.35&2497\\
&LDA-Sim+WNsyn&1.05&\bf 0.26&0.16&1715\\
&LDA-Sim+WNsynrel&1.20&\bf0.25&0.18&4984\\
&LDA-Sim+WNsr/hyp&1.47&0.24&0.33&1502\\
&LDA-Sim+WNsr/hyp+Ngrams&2.08 &0.23& \bf 0.42 &1929\\
&LDA-Sim+WNsr/hyp+Ngrams/l&\bf 2.46&\bf 0.25&\bf 0.43&1880\\

\hline
JRC&LDA unigram&1.42&0.24&0.53&807\\
&LDA+1000ngrams& 1.46&0.22&0.56&1140\\
&LDA-Sim+WNsyn&1.32&0.25&0.44&854\\
&LDA-Sim+WNsynrel&1.26&\bf 0.27&0.28&1367\\
&LDA-Sim+WNsynrel/hyp&\bf 1.57&0.24&0.54&823\\
&LDA-Sim+WNsr/hyp+Ngrams&1.54 &0.19&  0.64 &1093\\
&LDA-Sim+WNsr/hyp+Ngrams/l&\bf 1.58&0.18&\bf 0.68&1064\\
\hline
ACL&LDA unigram&1.63&0.24&0.51&1779\\
&LDA+1000ngrams&1.55&0.23&0.51&2277\\
&LDA-Sim+WNsyn&1.42&0.26&0.47&1853\\
&LDA-Sim+WNsynrel&1.26&0.27&0.35&2554\\
&LDA-Sim+WNsynrel/hyp&1.56&0.24&0.51&1785\\
&LDA-Sim+WNsr/hyp+Ngrams&2.72 &\bf 0.28&  0.69 &2164\\
&LDA-Sim+WNsr/hyp+Ngrams/l&\bf  3.04&\bf  0.28&\bf 0.76&2160\\
\hline
NIPS&LDA unigram&1.60&0.24&0.41&1284\\
&LDA+1000ngrams&1.54&0.24&0.41&1969\\
&LDA-Sim+WNsyn&1.34&0.26&0.39&1346\\
&LDA-Sim+WNsynrel&1.20&0.27&0.29&2594\\
&LDA-Sim+WNsynrel/hyp&1.78&0.25&0.43&1331\\
&LDA-Sim+WNsr/hyp+Ngrams&3.18 &\bf 0.31&  0.62&1740\\
&LDA-Sim+WNsr/hyp+Ngrams/l&\bf 3.27&\bf 0.30&\bf 0.67&1741\\

\hline
\end{tabular}
\end{center}
\caption{\label{WordNet} Integration of WordNet into topic models}
\end{table}
We add the Wordnet data in the following steps.
At the first step, we include WordNet synonyms (including multiword expressions) into the proposed similarity sets (LDA-Sim+WNsyn). At this step, frequencies of synonyms found in the same document are summed up in process LDA topic learning as described in Algorithm~\ref{algo:lda_sim_algorithm}. We can see that the kernel uniqueness becomes very low, topics are very close to each other in content (Table 2: LDA-Sim+WNsyn).
At the second step, we add word direct relatives (hyponyms, hypernyms, etc.) to similarity sets. Now the frequencies of semantically related words are added up enhancing the contribution into all topics of the current document.

\begin{table}[h]
\begin{center}
\begin{tabular}{|c|l|c|c|c|c|}
\hline 
\bf Collection& \bf Method& \bf TC-PMI  & \bf TC-NPMI & \bf Kernel Uniq&  \bf Perplex.\\
    \hline
Europarl&LDA unigram&1.20&0.24&0.33&1466\\
&LDA+1000ngram&1.19&0.23&0.35&2497\\
&LDA-Sim+EVsyn&1.57&0.24&0.43&1655\\
&LDA-Sim+EVsynrel&1.39&0.24&0.35&1473\\
&LDA-Sim+EVsr/hyp+Ngrams&\bf2.51&\bf 0.26&\bf0.50&1957\\
&LDA-Sim+EVsr/hyp+Ngrams/l&\bf2.5&\bf 0.25&0.45&1882\\
 \hline
JRC&LDA unigram&1.42&0.24&0.53&807\\
&LDA+1000ngrams& 1.46&0.22&0.56&1140\\
&LDA-Sim+EVsyn&1.65&\bf 0.25&0.57&857\\
&LDA-Sim+EVsynrel&1.71&\bf0.24&0.57&844\\
&LDA-Sim+EVsr/hyp+Ngrams&\bf1.91& 0.21&\bf 0.68&1094\\
&LDA-Sim+EVsr/hyp+Ngrams/l&1.5& 0.18&\bf 0.67&1061\\

\hline
\end{tabular}
\end{center}
\caption{\label{EuroVoc} Integration of EuroVoc into topic models }
\end{table}

The Table 2 shows that these two steps lead to great degradation of the topic model in most measures in comparison to the initial unigram model: uniqueness of kernels  abruptly decreases, perplexity at the second step grows by several times (Table 2: LDA-Sim+WNsynrel). It is evident that at this step the model has a poor quality. When we look at the topics, the cause of the problem seems to be clear. We can see the overgeneralization of the obtained topics. The topics are built around very general words such as "person", "organization", "year", etc. These words were initially frequent in the collection and then received additional frequencies from their frequent synonyms and related words.

Then we  suppose that these general words were used in texts to discuss specific events and objects, therefore, we change the constructions of the similarity sets in the following way:  we do not add word hyponyms  to its similarity set. Thus, hyponyms, which are usually more specific and concrete, should obtain additional frequencies from upper synsets and increase their contributions into the document topics. But the frequencies and contribution of hypernyms into the topic of the document are not changed.  And we see the great improvement of the model quality: the kernel uniqueness considerably improves, perplexity decreases to levels comparable with the unigram model, topic coherence characteristics also improve for most collections (Table 2:LDA-Sim+WNsynrel/hyp).

We further use the WordNet-based similarity sets with n-grams having the same components as described in \cite {Nokel2016}. All measures significantly improve for all collections (Table 2:LDA-Sim+WNsr/hyp+Ngrams).  At the last step, we try to apply the same approach to ngrams that was previously utilized to hyponym-hypernym relations: frequencies of shorter ngrams and words are summed to frequencies of longer ngrams but not vice versa. In this case we try to increase the contribution of more specific longer ngrams into topics. It can be seen (Table 2) that the kernel uniqueness grows significantly, at this step it is 1.3-1.6 times greater  than for the baseline  models  achieving 0.76 on the ACL collection (Table 2:LDA-Sim+WNsr/hyp+Ngrams/l).

At the second series of the experiments, we applied  EuroVoc information retrieval thesaurus to two European Union collections: Europarl and JRC.  In content, the EuroVoc thesaurus is much smaller than WordNet, it contains terms from economic  and political domains and does not include general abstract words. The results are shown in Table 3.
It can be seen that inclusion of EuroVoc synsets improves the topic coherence and increases kernel uniqueness (in contrast to results with WordNet). Adding ngrams further improves the topic coherence and kernel uniqueness.

At last we experimented with the Russian banking collection and utilized RuThes thesaurus. In this case we obtained improvement already on RuThes synsets and again adding ngrams further improved topic coherence and kernel uniqueness (Table 4). 

\begin{table}[h]
\begin{center}
\begin{tabular}{|c|l|c|c|c|c|}
\hline 
\bf Collection& \bf Processing  & \bf TC-PMI&\bf TC-NPMI & \bf Kernel Uniq&\bf Perplex.\\
    \hline

Banking &LDA unigram&1.81&0.29&0.54&1654\\
Collection& LDA+1000ngrams&2.01&0.30&0.60&2497\\
&LDA-Sim+RTsyn&2.03&0.29&0.63&2189\\
&LDA-Sim+RTsr/hyp+Ngrams&2.72&\bf 0.33 &\bf 0.70&2396 \\
&LDA-SIM+RTsr/hyp+Ngrams/l&\bf3.02& 0.31& 0.68&2311\\
  \hline

\end{tabular}
\end{center}
\caption{\label{RuThes} The results obtained for Russian Banking collection}
\end{table}

It is worth noting that adding ngrams sometimes worsens the TC-NPMI measure, especially on the JRC collection. This is due to the fact that in these evaluation frameworks, the topics' top elements contain a lot of multiword expressions, which rarely occur in Wikipedia, used for the coherence calculation, therefore the utilized automatic coherence  measures can  have insufficient evidence for correct estimates.

\section{Manual Evaluation of  Combined Topic Models}

To estimate the quality of topic models in a real task, we chose  Islam  informational portal "Golos Islama" (Islam Voice)\footnote {https://golosislama.com/} (in Russian). This portal contains both news articles related to Islam and  articles discussing Islam basics. We supposed that the thematic analysis of this specialized site can be significantly  improved with domain-specific knowledge described in the thesaurus form. We extracted the site contents using Open Web Spider\footnote {https://github.com/shen139/openwebspider/releases} and obtained 26,839  pages.

To combine knowledge with a topic model, we used RuThes thesaurus together with the additional block of the Islam thesaurus. The Islam thesaurus  contains more than 5 thousand Islam-related terms  including single words and expressions. 

For each combined model, we ran two experiments with 100 topics and with 200 topics. The generated topics were evaluated by two linguists, who had previously worked on the Islam thesaurus. The evaluation task was formulated as follows: the experts should read the top elements of the generated topics and try to formulate labels of these topics. The labels should be different for each topic in the set generated with a specific model. The experts should  also assign scores to the topics' labels:
\begin {itemize}
\item 2, if the label describes all or almost all elements of ten top elements of the topic
\item 1, if  the description is partial, that is, several elements do  not correspond to the label,
\item  0, if the label cannot be formulated.
\end {itemize}

Then we can sum up all the scores for each model under consideration and compare the total scores in value.  Thus, maximum values of the topic score are 200 for a 100-topic model and 400 for a 200-topic model. In this experiment we do not measure inter-annotator agreement for each  topic, but try to get expert's general impression.

\begin{table}[h]
\begin{center}
\begin{tabular}{|l|l|c|c|c|c|c|c|c|c|}
\hline 
\bf N&\bf Model & \multicolumn{4}{|c|}{\bf 100 topics}&\multicolumn{4}{|c|}{\bf 200 topics}\\
\hline 
&&Score&KernU&Prpl&RelC.&Score&KernU&Prpl&RelC.\\
\hline 

1       &LDA unigram    &163    &0.535  &2520&  0.05&   334&    0.507&  2169    &0.06\\
2       &LDA+1000phrases        &161&   0.569   &2901&  0.06    &316&   0.534   &2494   &0.06\\
3       &LDA+More10phrases&     148     &0.559& 3228&   0.05    &308&   0.527&  2774&   0.06\\
4       &LDA-Sim+
1000phrases     &180&   0.631&  2427&   0.13    &344&0.603&2044 &0.11\\
5&      LDA-Sim+More10phrases&  180&    0.615   &2886   &0.14   &337&   0.596&  2398    &0.12\\
6       &LDA-Sim+UnarySyn&      157&0.632       &1999&  0.17&   323     &0.587& 1707    &0.16\\
7&      LDA-Sim+
synrel+1000phrases       &159    &0.622& \bf 1797&       \bf 0.25&       301     &0.543& \bf 1577&\bf    0.27\\
8&      LDA-Sim+
synrel+
More10phrases   &150    &0.587  &2022   &\bf 0.26&      295     &0.526  &1758   &\bf 0.25\\
9&      LDA-Sim+
synrel/hyp+
1000phrases     &153    &0.656& 2163&\bf        0.26    &310&   0.603   &1900   &\bf 0.24\\
10      &LDA-Sim+
synrel/hyp+
More10phrases   &174&   0.636&  2476&\bf        0.24    &302&   0.244   &2476   &\bf 0.24\\
11&     LDA-Sim+
synrel/GL+
More10phrases   & \bf 186       & \bf 0.655     &\bf 1772&      \bf 0.25        &\bf 350        &\bf  0.612&    \bf 1464        &\bf 0.25\\
12&     LDA-Sim+
synrel/GL/hyp
        & \bf184        &\bf 0. 686     &2203   &\bf 0.24       &\bf 346        &\bf 0.644      &1812&  \bf 0.23\\
&+More10phrases&&&&&&&&\\

\hline 
\end{tabular}
\end{center}
\caption{\label{Labeling} Results of manual labeling of topic models for the Islam site}
\end{table}

Due to the complicated character of the Islam portal contents for automatic extraction (numerous words and names difficult for Russian morphological analyzers), we did not use automatic extraction of multiword expressions and exploited only phrases described in RuThes or in the Islam Thesaurus. We added thesaurus phrases in two ways: most frequent 1000 phrases (as in \cite {Lau2013,Nokel2016}) and phrases with frequency more than 10 (More10phrases):  the number of such phrases is 9351.

The results of the evaluation are shown in Table 5. The table contains the overall expert scores for a topic model (Score), kernel uniqueness as in the previous section (KernU), perplexity (Prpl).  Also for each model kernels, we calculated the average number of known relations between topics’s elements: thesaurus relations (synonyms and direct relations between concepts) and component-based relations between phrases (Relc).

It can be seen that if we add phrases without accounting component similarity (Runs 2, 3), the quality of topics decreases: the more phrases are added, the more the quality degrades. The human scores also confirm this fact. But if the similarity between phrase components is considered then the quality of topics significantly improves and becomes better than for unigram models (Runs 4, 5). All measures are better. Relational coherence between kernel elements also grows. The number of added phrases is not very essential.

Adding unary synonyms decreases the quality of the models (Run 6) according to human scores. But all other measures behave differently: kernel uniqueness is high, perplexity decreases, relational coherence grows. The problem of this model is in that non-topical, general words are grouped together, reinforce one another but do not look as related to any topic.
Adding all thesaurus relations is not very beneficial (Runs 7, 8). If we consider all relations except hyponyms, the human scores are better for corresponding runs (Runs 9, 10). Relational coherence in topics’ kernels achieves very high values: the quarter of all elements have some relations between each other, but it does not help to improve topics. The explanation is the same: general words can be grouped together.

\begin{table}[h!]
\begin{center}
\begin {footnotesize}
\begin{tabular}{|l|l|l|l|}
\hline
\bf N&\bf Unigram topic&\bf Phrase-enriched topic&\bf Thesaurus-enriched topic \\
\hline
& \bf Syria topic (Run 1)&\bf Syria topic (Run 5)&\bf Syria topic (Run 12)\\
& Relation coherence \bf 0.11&Relation coherence \bf 0.13&Relation coherence \bf 0.36\\
\hline
1.&\begin{otherlanguage*}{russian}сирия  \end{otherlanguage*} &\begin{otherlanguage*}{russian}сирия \end{otherlanguage*}&\begin{otherlanguage*}{russian}сирия \end{otherlanguage*}\\
&(Syria)&(Syria)&(Syria)\\
2.&\begin{otherlanguage*}{russian}сирийский  \end{otherlanguage*}&\begin{otherlanguage*}{russian}башар асад
\end{otherlanguage*}&\begin{otherlanguage*}{russian}сирийский
\end{otherlanguage*}\\
&(Syrian)&(Bashar al-Assad)&(Syrian)\\
3.&\begin{otherlanguage*}{russian}асад  \end{otherlanguage*}&\begin{otherlanguage*}{russian}сирийская оппозиция \end{otherlanguage*}&\begin{otherlanguage*}{russian}асад  \end{otherlanguage*}\\
&(Assad)&(Syrian opposition)&(Assad)\\
4.&\begin{otherlanguage*}{russian}оон \end{otherlanguage*} &\begin{otherlanguage*}{russian}сирийский \end{otherlanguage*}&\begin{otherlanguage*}{russian}дамаск \end{otherlanguage*}\\
&(UN)&(Syrian)&(Damask)\\
5.&\begin{otherlanguage*}{russian}оппозиция  \end{otherlanguage*}&\begin{otherlanguage*}{russian}режим асада \end{otherlanguage*}&\begin{otherlanguage*}{russian}башар асад \end{otherlanguage*}\\
&(opposition)&(al-Assad regime)&(Bashar al-Assad)\\
6.&\begin{otherlanguage*}{russian}башар      \end{otherlanguage*}&\begin{otherlanguage*}{russian}асад    \end{otherlanguage*}&\begin{otherlanguage*}{russian}сирийская оппозиция  \end{otherlanguage*}\\
&(Bashar)&(Assad)&(Syrian opposition)\\
7.&\begin{otherlanguage*}{russian}страна  \end{otherlanguage*}&\begin{otherlanguage*}{russian}сирийский режим \end{otherlanguage*}&\begin{otherlanguage*}{russian}оппозиция \end{otherlanguage*}\\
& (country)&(Syrian regime)&(opposition)\\
8.&\begin{otherlanguage*}{russian}дамаск  \end{otherlanguage*}&\begin{otherlanguage*}{russian}режим башара асада \end{otherlanguage*}&\begin{otherlanguage*}{russian}режим асада \end{otherlanguage*}\\
& (Damask)&(Bashar al-Assad regime)&(al-Assad regime)\\
9.&\begin{otherlanguage*}{russian}президент \end{otherlanguage*}&\begin{otherlanguage*}{russian}сирийская власть \end{otherlanguage*}&\begin{otherlanguage*}{russian}режим   \end{otherlanguage*}\\
&  (President)& (Syrian authorities )&(regime)\\

\hline
\hline
&\bf Orthodox church topic&\bf Orthodox church topic&\bf Orthodox church topic\\

& Relation coherence \bf 0.04&Relation coherence \bf 0.2&Relation coherence \bf 0.33\\
\hline
1.&\begin{otherlanguage*}{russian}православный \end{otherlanguage*} &\begin{otherlanguage*}{russian}русская православная цер- \end{otherlanguage*}&\begin{otherlanguage*}{russian}церковь \end{otherlanguage*}\\
&(orthodox)&\begin{otherlanguage*}{russian}ковь\end{otherlanguage*} (Russian orthodox church)&(church)\\
2.&\begin{otherlanguage*}{russian}церковь  \end{otherlanguage*}&\begin{otherlanguage*}{russian}православный \end{otherlanguage*}&\begin{otherlanguage*}{russian}православный
\end{otherlanguage*}\\
&(church)&(orthodox)&(orthodox)\\
3.&\begin{otherlanguage*}{russian}рпц  \end{otherlanguage*}&\begin{otherlanguage*}{russian}церковь  \end{otherlanguage*}&\begin{otherlanguage*}{russian}храм \end{otherlanguage*}\\
&(ROC, abbreviation)&(church)&(temple)\\
4.&\begin{otherlanguage*}{russian}патриарх \end{otherlanguage*} &\begin{otherlanguage*}{russian}русский язык \end{otherlanguage*}&\begin{otherlanguage*}{russian}православие \end{otherlanguage*} \\
&(patriarch)&(Russian language)&(orthodoxy)\\
5.&\begin{otherlanguage*}{russian}храм  \end{otherlanguage*}&\begin{otherlanguage*}{russian}рпц \end{otherlanguage*}&\begin{otherlanguage*}{russian}церковный \end{otherlanguage*}\\
&(temple)&(ROC, abbreviation)&(churchly)\\
6.&\begin{otherlanguage*}{russian}русский          \end{otherlanguage*}&\begin{otherlanguage*}{russian}православная церковь     \end{otherlanguage*}&\begin{otherlanguage*}{russian}русская православная цер-  \end{otherlanguage*}\\
&(Russian)&(orthodox church)&\begin{otherlanguage*}{russian}ковь  \end{otherlanguage*}(Russian orthodox church)\\
7.&\begin{otherlanguage*}{russian}московский  \end{otherlanguage*}&\begin{otherlanguage*}{russian}патриарх \end{otherlanguage*}&\begin{otherlanguage*}{russian}духовный \end{otherlanguage*}\\
&(Moscow)&(patriarch)&(spiritual)\\
8.&\begin{otherlanguage*}{russian}год \end{otherlanguage*}&\begin{otherlanguage*}{russian}кирилл \end{otherlanguage*}&\begin{otherlanguage*}{russian}русский \end{otherlanguage*}(russian)\\
& (year)&(Kirill)&(Russian)\\
9.&\begin{otherlanguage*}{russian}священник  \end{otherlanguage*}&\begin{otherlanguage*}{russian}государственный язык \end{otherlanguage*}&\begin{otherlanguage*}{russian}рпц \end{otherlanguage*}\\
& (priest)&(state language)&(ROC, abbr. for Russian  church)
\\
10.&\begin{otherlanguage*}{russian}кирилл  \end{otherlanguage*}&\begin{otherlanguage*}{russian}священник  \end{otherlanguage*}&\begin{otherlanguage*}{russian}собор \end{otherlanguage*}\\
& (Kirill, orthodox patriarch) &(priest)&(cathedral)\\
\hline
\end{tabular}
\end {footnotesize}
\end{center}
\caption{\label{TopicEx1} Comparison of similar topics in the unigram, phrase-based (Run 5) and the best thesaurus-enriched topic models (Run 12). } 
\end{table}

At last, we removed General Lexicon concepts from the RuThes data, which are top-level, non-thematic concepts that can be met in arbitrary domains \cite {Loukachevitch} and considered all-relations and without-hyponyms variants (Runs 11, 12). These last variants achieved maximal human scores because they add thematic knowledge and avoid general knowledge, which can distort topics. Kernel uniqueness is also maximal.

Table 6 shows  similar topics  obtained with the unigram, phrase-enriched  (Run 5) and the thesaurus-enriched topic model (Run 12). The Run-5 model adds thesaurus phrases with frequency more than 10 and accounts for the component similarity between phrases. The Run-12 model accounts both component relations and hypernym thesaurus relations.   All topics are of high quality, quite understandable. The experts evaluated them with the same high scores. 

Phrase-enriched and thesaurus-enriched topics convey the content using both single words and phrases. It can be seen that phrase-enriched topics contain more phrases.  Sometimes the phrases can create not very convincing relations such as  \textit {Russian church - Russian language}. It is explainable but does not seem much topical in this case.

The thesaurus topics seem  to convey the contents in the most concentrated way. In the Syrian  topic general word \textit {country} is absent; instead of \textit {UN} (United Nations), it contains word \textit {rebel}, which is closer to the Syrian situation.
In the Orthodox church topic, the unigram variant contains extra word \textit {year}, relations of words \textit {Moscow} and \textit {Kirill} to other words in the topic can be inferred only from the encyclopedic knowledge.

\section{Conclusion}

In this paper we presented  the approach for introducing  thesaurus information into  topic models. The main idea of the approach is based on the assumption that if related words or phrases co-occur in the same text,  their frequencies should be enhanced and this action leads to their mutual larger contribution into topics found in this text. 

In the experiments on four English collections, it was shown that the direct implementation of this idea using WordNet synonyms and/or direct relations leads to great degradation of the unigram model. But the correction of initial assumptions and excluding  hyponyms from frequencies adding  improve  the model and makes it much better than the initial model in several measures. Adding ngrams in a similar manner  further improves the model. 

Introducing information from domain-specific thesaurus EuroVoc led to improving the initial model without the additional assumption, which can be explained by the absence of general abstract words in such information-retrieval thesauri. 

We also considered  thematic analysis of an Islam Internet site and evaluated the combined topic models manually. We found that the best, understandable topics are obtained by adding domain-specific thesaurus knowledge (domain terms, synonyms, and relations).

\subsubsection*{Acknowledgments.} This study is supported by Russian Scientific Foundation in part concerning the combined approach uniting thesaurus information and probabilistic topic models (project N16-18-02074).  The study on application of the approach to content analysis of Islam sites is supported by Russian Foundation for Basic Research (project N 16-29-09606).


\end{document}